\documentclass[11pt]{article}

\usepackage[margin=1in]{geometry}
\usepackage{amsmath,amssymb}
\usepackage{booktabs}
\usepackage{graphicx}
\usepackage{microtype}
\usepackage{enumitem}
\usepackage{natbib}
\usepackage{xcolor}
\usepackage{hyperref}
\hypersetup{
  colorlinks=true,
  linkcolor=blue!60!black,
  citecolor=blue!60!black,
  urlcolor=blue!60!black
}
\graphicspath{{figures/}}

\newcommand{\FSI}{\mathrm{FSI}}
\newcommand{\PSI}{\mathrm{PSI}}
\newcommand{\nFSI}{\mathrm{nFSI}}
\newcommand{\tokspread}{\Delta_{\mathrm{tok}}}

\title{Format Sensitivity Index: Token-Controlled Prompt Wrapper Robustness and Schema Compliance in LLM Benchmarking}
\author{Deep Pankajbhai Mehta\\
Adobe Inc.}
\date{}

\begin{document}
\maketitle

\begin{abstract}
Prompt wrappers often differ only in formatting, yet they can change model scores enough to flip leaderboard conclusions. We study this variance under a token-controlled protocol and introduce two complementary metrics: the Format Sensitivity Index (FSI), the accuracy range induced by wrapper choice, and the Parseability Sensitivity Index (PSI), the corresponding range in answer parseability. Across 140,000 OpenRouter generations spanning 7 question-answering tasks, 5 wrapper families, and 4 instruct models from 7B to 72B parameters, we find that mean FSI varies by more than 30 times across models and is largely explained by compliance failures. A fixed-effects regression shows that parseability remains a strong predictor of accuracy even after controlling for task, model, and wrapper. We argue that reporting accuracy without wrapper variance and compliance is statistically fragile, and we provide practical recommendations for both benchmarking and structured-output deployments.
\end{abstract}

\section{Introduction}

Benchmarks increasingly evaluate large language models (LLMs) through prompts that include a wrapper: delimiters, JSON instructions, step-by-step templates, and similar formatting that is meant to be semantically irrelevant. In practice, wrappers are rarely standardized across papers, toolkits, or vendors. This creates a methodological failure mode: a model can look strong or weak depending on the wrapper rather than the underlying capability.

Our experiments expose an extreme example. On the same set of 7 tasks, the same model can score near random accuracy under a strict JSON wrapper while exceeding 0.75 accuracy under a delimiter-based structured wrapper. The difference is not subtle prompting artistry. It is frequently a compliance problem: if the output is not parseable by the benchmark's answer extractor, the run is scored as incorrect.

Prior work has documented sensitivity to prompt formatting in few-shot settings \citep{sclar2024formatting,chatterjee2024posix}. However, most benchmarking pipelines still report a single number per model per task and often ignore structured-output failure modes that dominate real deployments, including tool calls, schema adherence, and database queries. We contribute a simple, reproducible way to quantify wrapper-driven variance in a setting that resembles common evaluation practice.

\paragraph{Contributions.}
\begin{itemize}[leftmargin=*]
  \item \textbf{Two sensitivity metrics.} We define FSI and PSI as wrapper-induced ranges of accuracy and parseability, and we report bootstrap confidence intervals plus a normalized sensitivity variant to reduce dependence on mean accuracy.
  \item \textbf{Token-controlled evaluation.} We implement a protocol that pads prompts to a fixed character budget, logs realized prompt token counts, and quantifies residual token spread.
  \item \textbf{Large empirical study.} We evaluate 4 instruct models across 7 tasks and 5 wrapper families, yielding 140,000 generations, and show that format sensitivity differs sharply across models.
  \item \textbf{Mechanism analysis.} We show that parseability strongly mediates accuracy differences, supported by correlation and fixed-effects regression.
\end{itemize}

\section{Related Work}

\paragraph{Prompt-format sensitivity.}
LLMs can be brittle to meaning-preserving prompt changes, including formatting \citep{sclar2024formatting,chatterjee2024posix}. PromptBench studies adversarial prompt robustness \citep{zhu2023promptbench}, and surveys catalog broader evaluation risks \citep{chang2023survey}. Our setting is narrower but common: instruction-following, single-turn QA, and wrappers that aim to change only output format.

\paragraph{Benchmarking methodology.}
Evaluation frameworks such as HELM emphasize standardization and transparency \citep{liang2022helm}. Toolkits such as \texttt{lm-evaluation-harness} provide a shared interface for benchmark execution \citep{eleutherai2024lmeval}. Our results suggest that wrapper choice is an additional axis of variance that should be tracked alongside decoding settings and dataset preprocessing.

\paragraph{Structured outputs and constrained decoding.}
Production systems increasingly require schema adherence. Prompt-based JSON instructions are brittle, motivating constrained decoding and grammar methods \citep{geng2023grammar,ugare2024syncode,park2024grammar,poesia2022synchromesh}. API providers also expose structured-output modes that enforce JSON validity or schema constraints at decoding time \citep{openai2024structured,anthropic2024tooluse}. Our work complements these efforts by measuring how much score variance arises when structure is requested but not enforced.

\section{Experimental Setup}

\subsection{Tasks and Models}

We evaluate 7 widely used QA-style benchmarks spanning Boolean, multiple-choice, and numeric answers: BoolQ \citep{clark2019boolq}, PIQA \citep{bisk2020piqa}, ARC-Challenge \citep{clark2018arc}, HellaSwag \citep{zellers2019hellaswag}, WinoGrande \citep{sakaguchi2020winogrande}, CommonsenseQA \citep{talmor2019commonsenseqa}, and GSM8K \citep{cobbe2021gsm8k}. For each task we sample 200 examples.

We test four instruct-tuned models accessed via OpenRouter: Mistral-7B-Instruct-v0.3 \citep{jiang2023mistral}, Gemma-2-9B-IT \citep{gemma2024gemma2}, Phi-4 \citep{abdin2024phi4}, and Qwen-2.5-72B-Instruct \citep{qwen2024qwen25}. This set intentionally mixes scales from 7B to 72B. We treat scale as a confound in our conclusions and discuss it in Section~\ref{sec:limitations}.

\subsection{Prompt Wrappers}

A wrapper is a template that surrounds the same task content, including the question and options, but changes formatting and output constraints. We evaluate five wrapper families: (1) \texttt{plain}, with minimal instructions; (2) \texttt{json}, strict JSON with an \texttt{answer} field; (3) \texttt{structured}, key-value fields; (4) \texttt{structured\_delim}, key-value fields plus explicit delimiter tokens; and (5) \texttt{deliberate}, a scratchpad plus final answer field. Representative excerpts are shown in Table~\ref{tab:wrappers}; full templates are summarized in Appendix~\ref{app:templates}.

\begin{table}[t]
\centering
\small
\caption{Representative wrapper excerpts. The task content is held fixed across wrappers. We parse only the final answer field.}
\label{tab:wrappers}
\begin{tabular}{p{0.42\linewidth}p{0.42\linewidth}}
\toprule
\textbf{JSON} & \textbf{Structured with delimiters} \\
\midrule
\texttt{Return a JSON object:}\newline
\texttt{\{"answer": ...\}}\newline
\texttt{No extra keys. No markdown.}
&
\texttt{<BEGIN ANSWER>}\newline
\texttt{answer: ...}\newline
\texttt{<END ANSWER>} \\
\bottomrule
\end{tabular}
\end{table}

\subsection{Token-Controlled Protocol}

Wrappers differ in surface length, which changes prompt token counts and can affect accuracy. To reduce this confound, we pad each prompt to a fixed character budget per task by inserting whitespace in wrapper-specific padding slots. We then log realized prompt tokens for every request. For each task, model, example, and seed, we compute the relative prompt-token spread across wrappers:
\begin{equation}
\tokspread = \frac{\max_f T_f - \min_f T_f}{\frac{1}{|F|}\sum_f T_f},
\label{eq:tokspread}
\end{equation}
where $T_f$ is prompt tokens under wrapper $f$. This protocol does not perfectly equalize tokens because tokenizers segment whitespace differently. We therefore report results on the full set and on a matched subset filtered by $\tokspread \leq 5\%$.

\subsection{Decoding and Sampling}

For each model, task, and wrapper, we generate answers for 200 examples using 5 independent seeds, yielding 1,000 generations per cell and 140 cells overall. We set \texttt{max\_tokens=24} and record completion lengths. Other decoding parameters are held constant across wrappers. Across the dataset, 42.3\% of task, model, wrapper, and example groups produce more than one distinct prediction over the five seeds, so we treat seeds as stochastic replicates.

\subsection{Answer Extraction, Accuracy, and Parseability}

We use task-aware extraction that maps each model output to a canonical answer string. Outputs that cannot be mapped are marked unparseable and scored as incorrect. Accuracy is the mean of the 0/1 correctness indicator over all generations. We additionally compute parseability as the fraction of generations with a valid extracted answer. Appendix~\ref{app:parsing} reports strict versus normalized extraction variants.

\section{Sensitivity Metrics}

Let $A_{m,t,f}$ denote accuracy for model $m$ on task $t$ under wrapper $f$, and let $P_{m,t,f}$ denote parseability. We define:
\begin{align}
\FSI(m,t) &= \max_f A_{m,t,f} - \min_f A_{m,t,f}, \\
\PSI(m,t) &= \max_f P_{m,t,f} - \min_f P_{m,t,f}.
\end{align}
FSI is a range statistic and can be sensitive to noise. We therefore report bootstrap confidence intervals and a normalized variant:
\begin{equation}
\nFSI(m,t) = \frac{\FSI(m,t)}{\frac{1}{|F|}\sum_f A_{m,t,f} + 10^{-6}}.
\label{eq:nfsi}
\end{equation}

\section{Results}

\subsection{Format Sensitivity Varies Sharply Across Models}

\begin{figure}[t]
\centering
\includegraphics[width=0.82\linewidth]{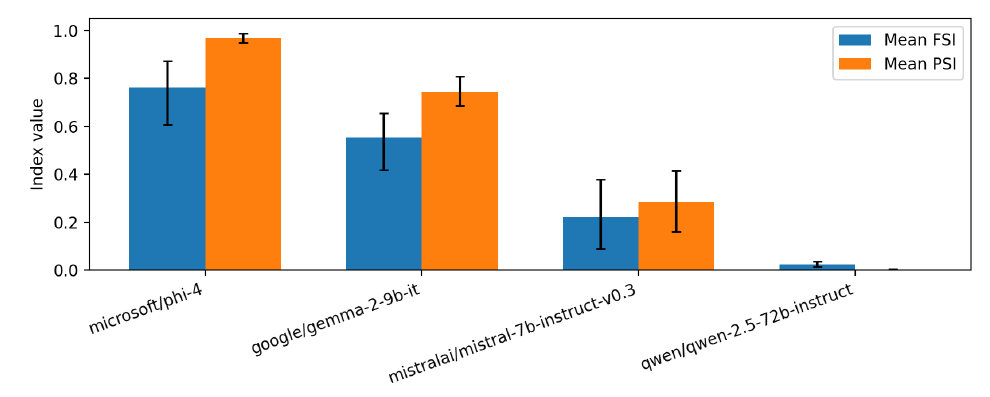}
\caption{Mean FSI and PSI across tasks, with bootstrap confidence intervals over tasks from 10,000 resamples. Models differ by more than 30 times in mean FSI.}
\label{fig:fsi_psi_summary}
\end{figure}

Figure~\ref{fig:fsi_psi_summary} summarizes mean FSI and PSI across tasks. Qwen-2.5-72B is nearly format-invariant, with mean FSI 0.024 and 95\% CI [0.014, 0.035], while Phi-4 is highly sensitive, with mean FSI 0.763 and CI [0.607, 0.872]. Normalized sensitivity accentuates this contrast: Phi-4 has mean nFSI 2.37 and CI [2.19, 2.63], indicating that wrapper choice can dominate the mean score. Table~\ref{tab:sensitivity} reports all values.

\begin{table}[t]
\centering
\small
\caption{Sensitivity summary by model, macro averaged over 7 tasks. We report mean FSI, PSI, and nFSI with 95\% bootstrap confidence intervals over tasks.}
\label{tab:sensitivity}
\begin{tabular}{lccc}
\toprule
\textbf{Model} & \textbf{FSI} & \textbf{PSI} & \textbf{nFSI} \\
\midrule
Phi-4 & 0.763 [0.607, 0.872] & 0.969 [0.948, 0.988] & 2.370 [2.188, 2.632] \\
Gemma-2-9B & 0.554 [0.418, 0.655] & 0.743 [0.685, 0.807] & 0.926 [0.856, 1.004] \\
Mistral-7B & 0.221 [0.088, 0.379] & 0.284 [0.165, 0.417] & 0.601 [0.337, 0.891] \\
Qwen-2.5-72B & 0.024 [0.014, 0.035] & 0.003 [0.001, 0.005] & 0.037 [0.016, 0.068] \\
\bottomrule
\end{tabular}
\end{table}

\subsection{Parseability Mediates a Large Fraction of Wrapper Variance}

\begin{figure}[t]
\centering
\includegraphics[width=0.72\linewidth]{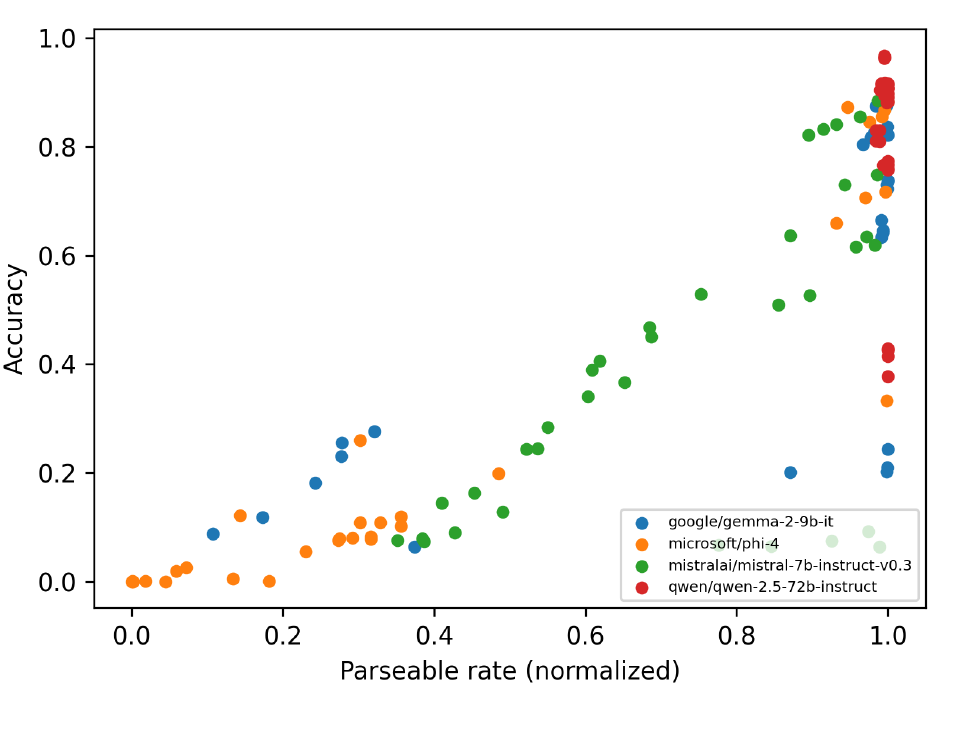}
\caption{Accuracy versus parseability across all model, task, and wrapper cells. Lower parseability often implies near-zero accuracy.}
\label{fig:accuracy_parseability}
\end{figure}

Across the 140 model, task, and wrapper cells, accuracy and parseability are strongly correlated with Pearson $r=0.825$. Figure~\ref{fig:accuracy_parseability} shows that low accuracy frequently occurs when parseability collapses. This is especially pronounced for wrappers that request rigid structure. For example, under the JSON wrapper, Phi-4 produces outputs beginning with a markdown code fence in 85.3\% of generations, yielding parseability 2.8\% and accuracy 0.7\%.

Correlation alone does not establish mediation. We therefore fit a weighted least-squares regression at the cell level,
\begin{equation}
A_{m,t,f} = \beta P_{m,t,f} + \alpha_m + \gamma_t + \delta_f + \epsilon,
\label{eq:regression}
\end{equation}
with fixed effects for model $m$, task $t$, and wrapper $f$. Parseability remains a strong predictor, with $\hat{\beta}=0.819\pm0.034$, explaining substantial residual variance with partial $R^2=0.82$. This supports a practical claim: many wrapper-induced score changes are driven by whether the output is machine-consumable.

\subsection{Which Wrappers Win Most Often}

\begin{figure}[t]
\centering
\includegraphics[width=0.66\linewidth]{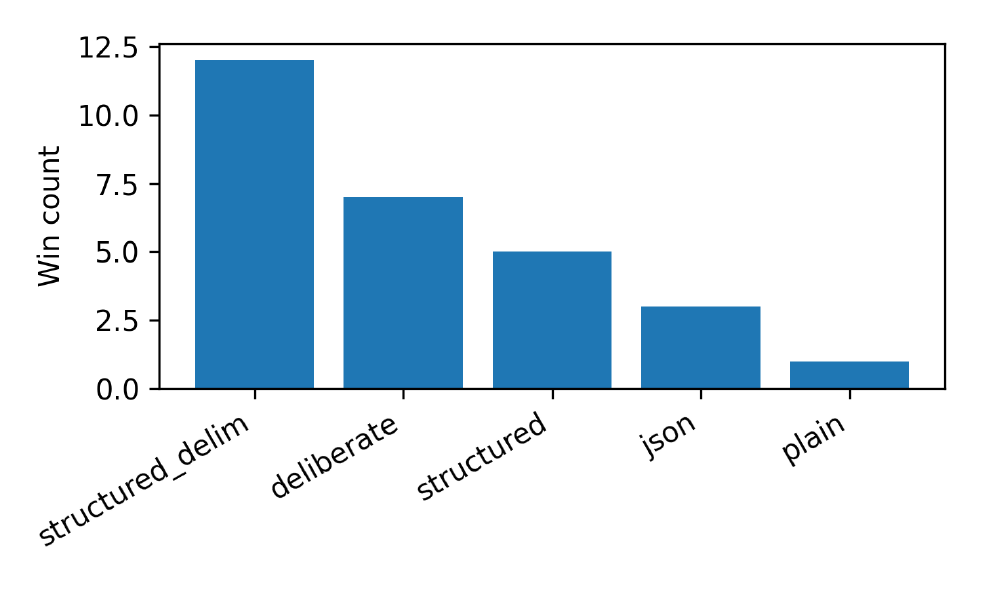}
\caption{Wrapper win counts over 28 model and task combinations.}
\label{fig:wrapper_wins}
\end{figure}

Wrapper choice is not a pure compliance tradeoff. Some wrappers raise accuracy even when parseability is already high, suggesting that formatting can also influence reasoning. Figure~\ref{fig:wrapper_wins} counts, for each model and task, which wrapper achieves the highest accuracy. The delimiter-based structured wrapper wins most frequently, 12 out of 28 combinations, but deliberate and structured wrappers also win nontrivially. This supports reporting a distribution over wrappers rather than selecting a single best format.

\subsection{Token Control and Residual Token Spread}

\begin{figure}[t]
\centering
\includegraphics[width=0.72\linewidth]{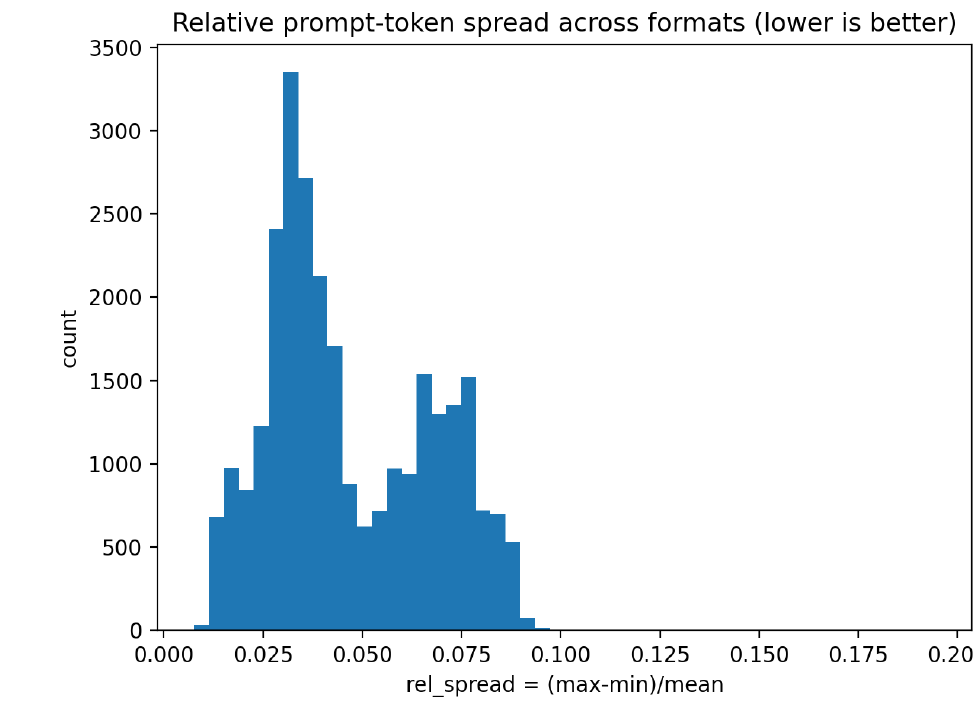}
\caption{Distribution of prompt-token spread $\tokspread$ across wrappers, after padding to equal character budgets.}
\label{fig:token_spread}
\end{figure}

Our padding protocol reduces, but does not eliminate, prompt token differences. Across all task, model, example, and seed groups, the median relative token spread is 4.1\%, and 61.2\% of groups satisfy $\tokspread \leq 5\%$. Figure~\ref{fig:token_spread} shows the spread distribution. Crucially, conclusions are stable on the matched subset: mean FSI is 0.391 on the full set and 0.394 after filtering by $\tokspread \leq 5\%$.

\section{Discussion and Recommendations}

\paragraph{Benchmark reporting.}
If wrapper choice can change accuracy by 0.76, then point estimates without wrapper variance invite overinterpretation. We recommend that benchmarking papers and leaderboards report (i) the wrapper family used, (ii) prompt token counts, (iii) parseability rates, and (iv) an uncertainty-aware sensitivity metric such as FSI with confidence intervals. A simple alternative is to report the interval $[\min_f A_{m,t,f}, \max_f A_{m,t,f}]$ alongside the mean.

\paragraph{Structured-output deployments.}
Prompting alone is often insufficient when strict schemas are required. If the downstream consumer is brittle, then parseability collapse can dominate measured accuracy. Decoder-time constraints, including grammars, schema-constrained decoding, or vendor structured-output modes, can reduce PSI by construction \citep{geng2023grammar,openai2024structured}. We view FSI and PSI as diagnostics for when such enforcement is necessary.

\section{Limitations}
\label{sec:limitations}

First, FSI is a range statistic and can be influenced by outliers and sampling noise. We mitigate this by reporting bootstrap confidence intervals and nFSI, but richer uncertainty modeling, such as hierarchical bootstraps over examples and tasks, is future work. Second, our token control pads by characters and logs prompt tokens, but it does not strictly equalize token budgets. A stricter control could truncate or pad to equal token length per wrapper. Third, our model set is small and mixes scales, so robustness may reflect size as well as architecture. Fourth, we focus on single-turn QA tasks; multi-turn tool use and code generation may show different compliance dynamics. Finally, parser strictness matters: we report strict and normalized extraction variants, but we do not benchmark decoding-time enforcement or vendor JSON modes directly.

\section{Conclusion}

Prompt wrappers are an underreported confound in LLM benchmarking. FSI and PSI quantify how much a model's score depends on wrapper choice and whether that dependence is driven by schema compliance. In our study, sensitivity ranges from near zero to dominating the mean score. We argue that reliable benchmarking and reliable structured-output systems both require measuring, and often reducing, format sensitivity.

\section*{Impact Statement}

This paper is about evaluation methodology. More transparent reporting of wrapper variance and compliance can improve reproducibility and reduce misleading comparisons. However, sensitivity metrics might also be used to search for adversarial wrappers that inflate scores. We therefore recommend multi-wrapper reporting and public release of templates so that results can be audited.

\appendix

\section{Full Wrapper Templates}
\label{app:templates}

We summarize the output constraints used by each wrapper family in Table~\ref{tab:full_wrappers}. The same task content is held fixed across wrappers.

\begin{table}[h]
\centering
\small
\caption{Wrapper families and required output forms.}
\label{tab:full_wrappers}
\begin{tabular}{lp{0.65\linewidth}}
\toprule
\textbf{Wrapper} & \textbf{Output specification} \\
\midrule
\texttt{plain} & Output only the final answer. \\
\texttt{json} & Output a single JSON object with exactly one key, \texttt{answer}. No markdown. \\
\texttt{structured} & Output key-value lines including \texttt{answer: ...}. \\
\texttt{structured\_delim} & Output only a \texttt{<BEGIN ANSWER>} $\ldots$ \texttt{<END ANSWER>} block containing \texttt{answer: ...}. \\
\texttt{deliberate} & Output reasoning, then a final line \texttt{answer: ...}. \\
\bottomrule
\end{tabular}
\end{table}

\section{Parsing Variants and Seed Aggregation}
\label{app:parsing}

\paragraph{Normalized extraction.}
The main paper reports a task-aware extraction that tolerates minor formatting differences: for multiple-choice tasks we extract the first letter in \{A, B, C, D, E\}; for booleans we map to \{true, false\}; and for numeric answers we extract the last number.

\paragraph{Strict extraction.}
We also compute a strict validity rate that requires an exact canonical token. Results and correlations are qualitatively similar.

\paragraph{Seed aggregation.}
Main-text accuracy pools all example and seed generations. An alternative is to compute a per-example majority vote over the five seeds, then average over examples. Across the 140 cells, majority-vote accuracy is on average 0.008 lower; the largest absolute deviation is 0.121.

\section{Additional Plots}

\begin{figure}[p]
\centering
\includegraphics[width=\linewidth]{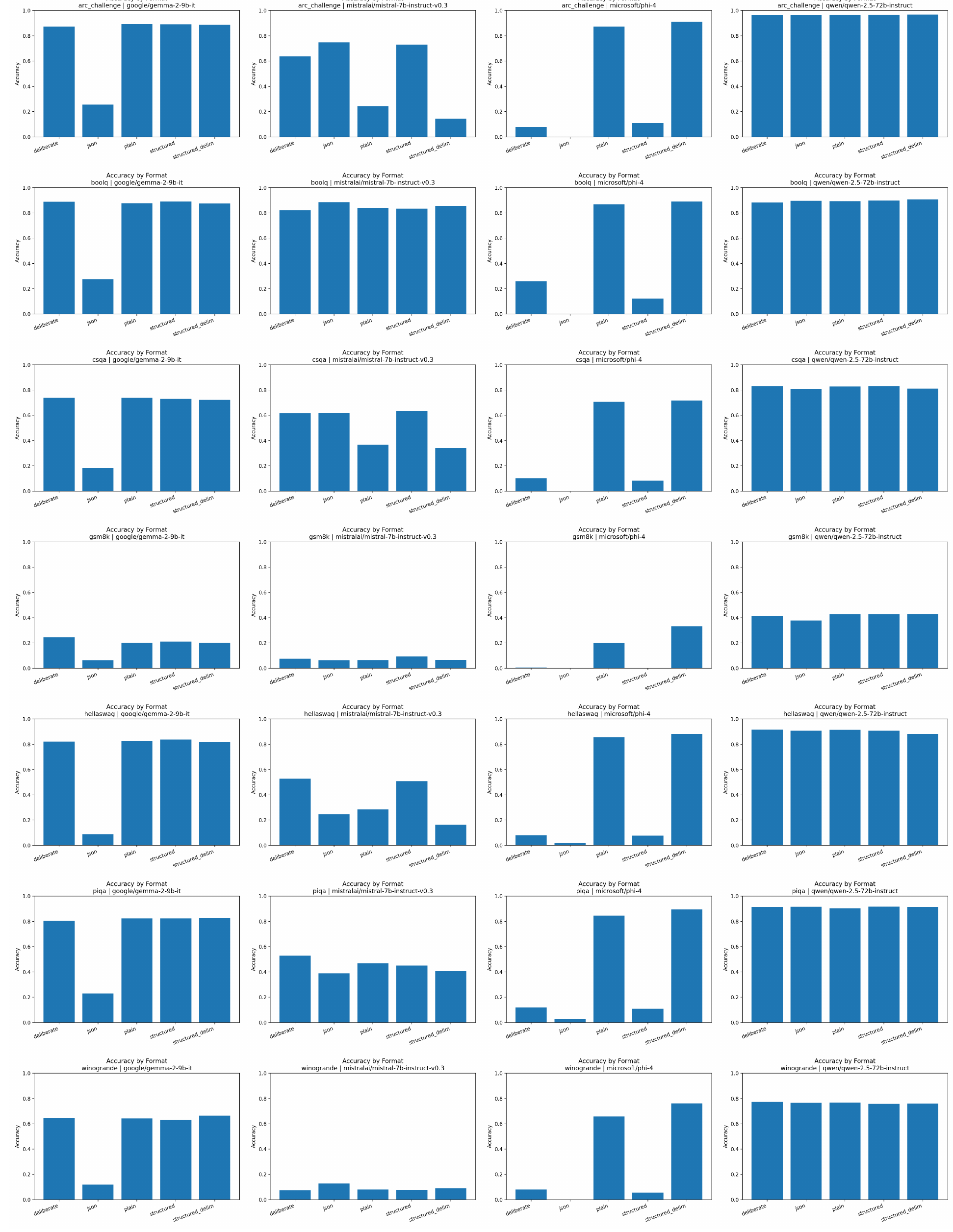}
\caption{Accuracy by model, wrapper, and task.}
\label{fig:accuracy_grid}
\end{figure}

\begin{figure}[p]
\centering
\includegraphics[width=\linewidth]{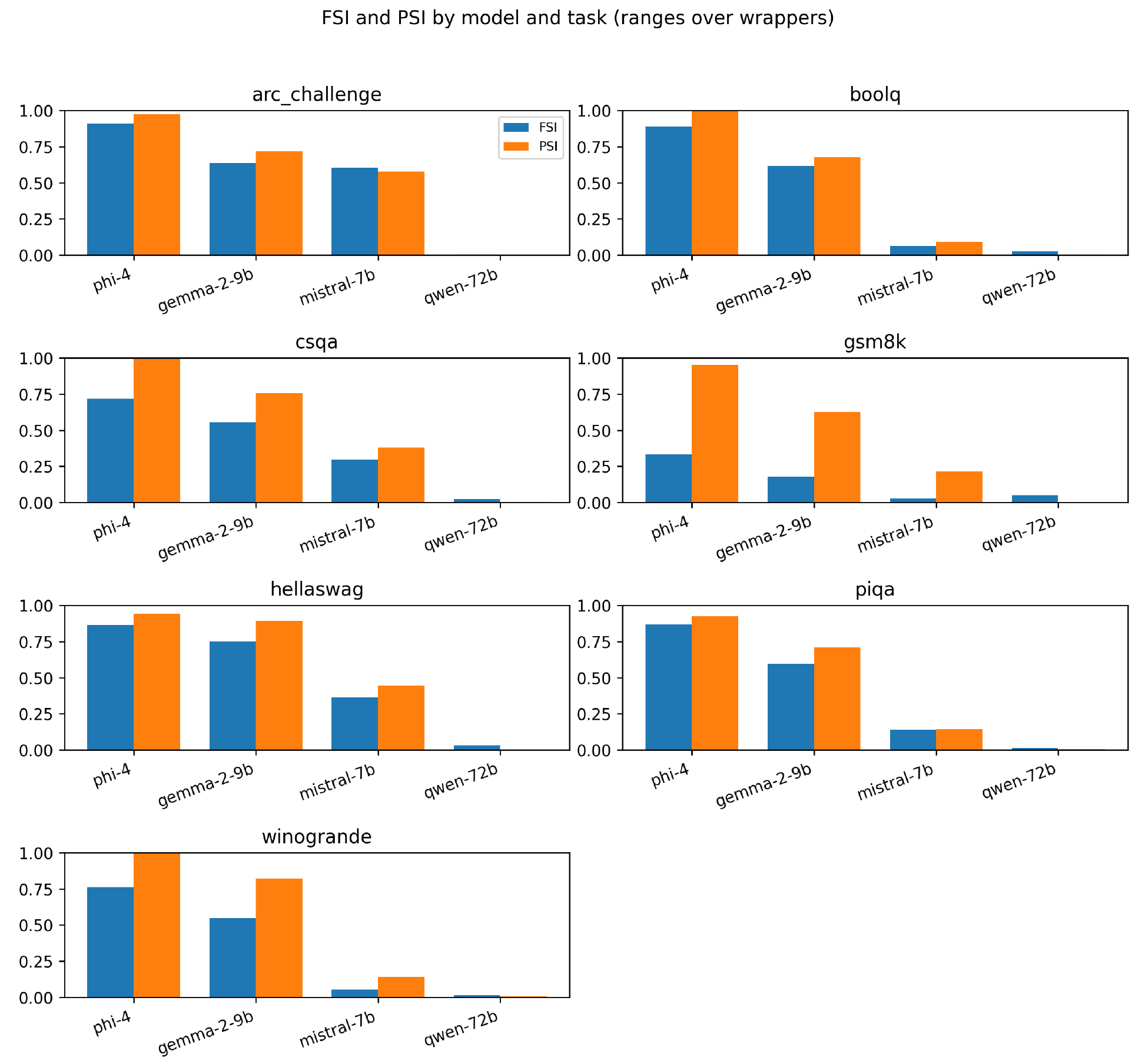}
\caption{FSI and PSI by model and task. Points are wrapper ranges.}
\label{fig:fsi_psi_by_task}
\end{figure}

Figures~\ref{fig:accuracy_grid} and~\ref{fig:fsi_psi_by_task} provide per-task accuracy and sensitivity visualizations.

\end{document}